\def\year{2020}\relax
\documentclass[letterpaper]{article} 
\usepackage{aaai20}  

\usepackage{subcaption}
\usepackage{times}  
\usepackage{helvet} 
\usepackage{courier}  
\usepackage[hyphens]{url}  
\usepackage{graphicx} 
\usepackage{graphicx}  
\usepackage[utf8]{inputenc} 
\usepackage[T1]{fontenc}    
\usepackage{hyperref}       
\usepackage{url}            
\usepackage{booktabs}       
\usepackage{amsfonts}       
\usepackage{nicefrac}       
\usepackage{microtype}      
\usepackage{siunitx}
\newcommand{\ra}[1]{\renewcommand{\arraystretch}{#1}}
\usepackage{graphicx}
\usepackage{amsmath}
\usepackage{amssymb}
\usepackage{amsthm}
\usepackage{bbold}
\usepackage{nth}
\usepackage[round]{natbib}
\usepackage[symbol]{footmisc}
\newcommand*\samethanks[1][\value{footnote}]{\footnotemark[#1]}

\title{Video Game Level Repair via Mixed Integer Linear Programming}

\author{
Hejia Zhang\textsuperscript{\rm 1}\thanks{Denotes equal contribution.},
Matthew C. Fontaine\textsuperscript{\rm 1}\samethanks{},
Amy K. Hoover\textsuperscript{\rm 3}, \\
\Large \textbf{Julian Togelius\textsuperscript{\rm 2},
Bistra Dilkina\textsuperscript{\rm 1},
Stefanos Nikolaidis\textsuperscript{\rm 1}}\\ 
\textsuperscript{\rm 1}Viterbi School of Engineering, University of Southern California, \{hejiazha,mfontain,dilkina,nikolaid\}@usc.edu\\ %
\textsuperscript{\rm 2}Tandon School of Engineering, New York University, julian@togelius.com \\
\textsuperscript{\rm 3}Ying Wu College of Computing, New Jersey Institute of Technology, ahoover@njit.edu 
}
\usepackage{xcolor}

\usepackage[normalem]{ulem}

\begin{document}

\maketitle

\begin{abstract}
Recent advancements in procedural content generation via machine learning enable the generation of video-game levels that are aesthetically similar to human-authored examples. However, the generated levels are often unplayable without additional editing. We propose a ``generate-then-repair'' framework for automatic generation of playable levels adhering to specific styles. The framework constructs levels using a generative adversarial network (GAN) trained with human-authored examples and repairs them using a mixed-integer linear program (MIP) with playability constraints. A key component of the framework is computing minimum cost edits between the GAN generated level and the solution of the MIP solver, which we cast as a minimum cost network flow problem. Results show that the proposed framework generates a diverse range of playable levels, that capture the spatial relationships between objects exhibited in the human-authored levels.\footnote{Code of the experiments is available at: \url{https://github.com/icaros-usc/milp_constrained_gan}}
\end{abstract}

\section{Introduction}

We focus on the problem of automatically generating video game levels that are aesthetically similar to human-authored examples, while satisfying playability constraints. For example, we would like to generate a variety of different \textit{Zelda} levels; on one hand, we require these levels to be playable; they should include a character that the user can control, a door to exit the level, and a key to open the door. At the same time, the levels should have an aesthetic appeal; a level with a character, a key, and a door next to each other is technically playable, but does not have the appeal of a level created by a human designer. 





\begin{figure}[t!]
\includegraphics[draft=false,width=1.0\columnwidth]{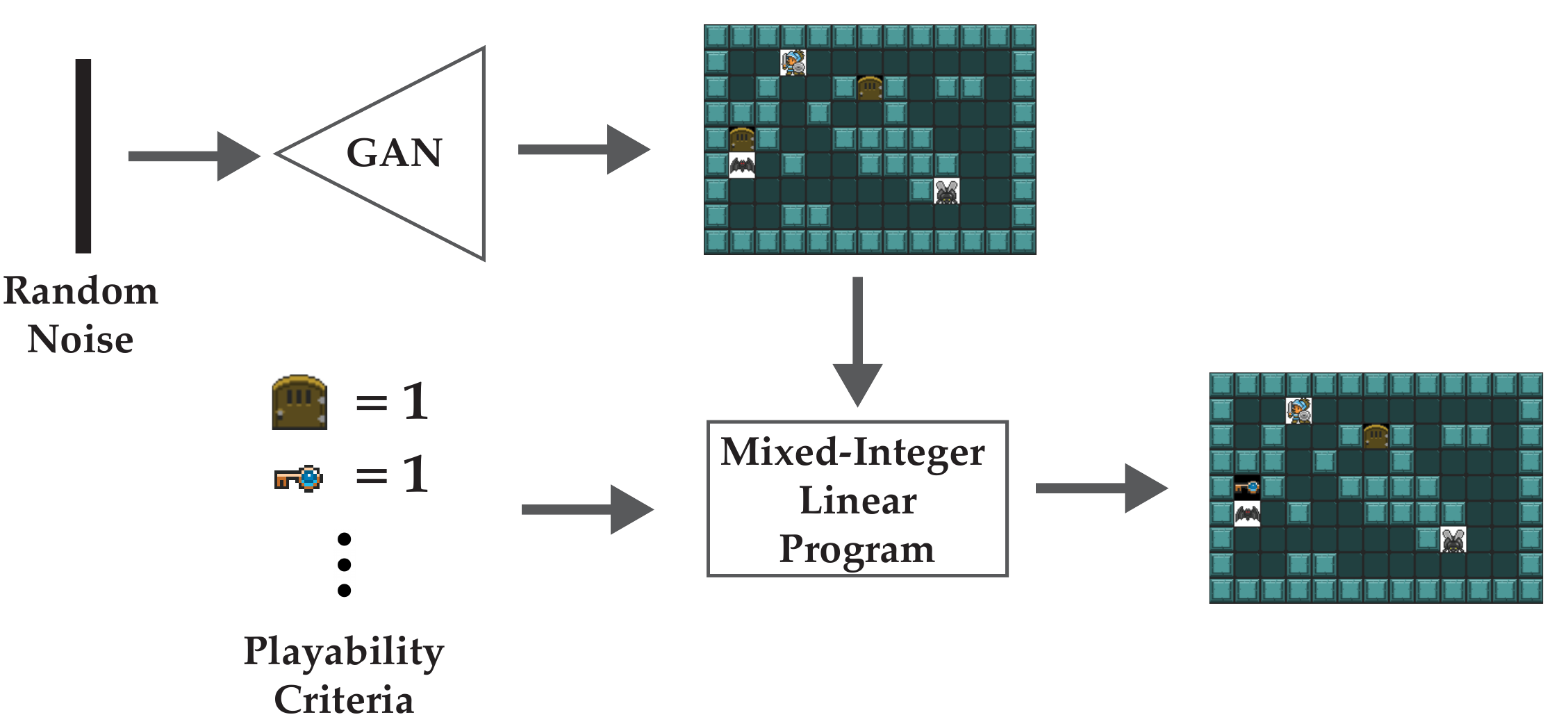}
\caption{An overview of our framework for generating aesthetically pleasing, playable game levels by using a mixed-integer linear program to repair GAN-generated levels.}
\label{fig:overview}
\end{figure}

Addressing the research question of procedurally generating aesthetically appealing video game levels that satisfy playability constraints is challenging. Using machine learning methods, level generators can be trained on existing levels so as to learn to reproduce aspects of their style~\citep{summerville:tog18}. 
In particular, recent advancements in generative adversarial networks (GANs) enable the creation of levels that are stylistically similar to human examples~\citep{volz:gecco18, giacomello:gem18}. At the same time, many levels generated this way are not playable. \citet{torrado:cog20} demonstrated that GANs frequently fail to encode playability criteria.

One approach to adhering to playability constraints is to encode the search space of possible levels via constraint programming (CP)~\citep{smith:tanagra11} or answer set programming (ASP)~\citep{smith:coai11} and then use a search algorithm to find a feasible solution. It is not clear how to combine such methods with machine learning to reproduce a given style. 

Rather than encoding constraints in the level generation process, we propose a \textit{generate-then-repair} approach for first generating levels from models trained on human-authored examples and then repairing them with minimum cost edits to render them playable.

\begin{figure*}[!t]
\centering

\begin{tabular}{lccccc}
\begin{subfigure}[l]{.08\linewidth}
\begin{flushleft}    \hspace{\bibindent}\raisebox{\dimexpr 2.5cm-\height}{  
    Human
    }        \end{flushleft}

    \end{subfigure}
&
\begin{subfigure}[b]{.17\linewidth}
\centering
  \begin{tabular}{cc}
 \resizebox{1.0\linewidth}{!}{
  \includegraphics[width=\linewidth]{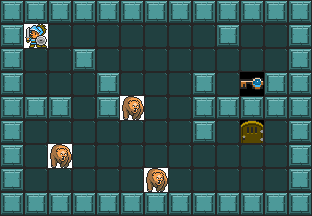}
   }
   \end{tabular}
 \label{fig:plotT0}
\end{subfigure}
\begin{subfigure}[b]{.17\linewidth}
\centering
  \begin{tabular}{cc}
  \resizebox{1.0\linewidth}{!}{
  \includegraphics[width=\linewidth]{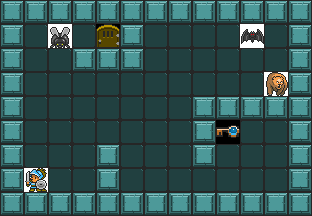}
   }
   \end{tabular}
 \label{fig:plotT0}
\end{subfigure}
\begin{subfigure}[b]{.17\linewidth}
\centering
  \begin{tabular}{cc}
    \resizebox{1.0\linewidth}{!}{
  \includegraphics[width=\linewidth]{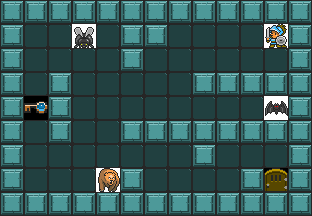}
   }  
   \end{tabular}
 \label{fig:plotT0}
\end{subfigure}
\begin{subfigure}[b]{.17\linewidth}
\centering
  \begin{tabular}{cc}
    \resizebox{1.0\linewidth}{!}{
  \includegraphics[width=\linewidth]{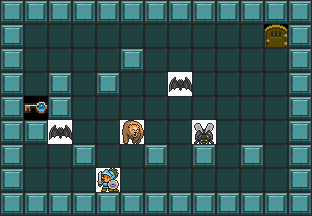}
   }
   \end{tabular}
 \label{fig:plotT0}
\end{subfigure}
\begin{subfigure}[b]{.17\linewidth}
\centering
  \begin{tabular}{cc}
    \resizebox{1.0\linewidth}{!}{
  \includegraphics[width=\linewidth]{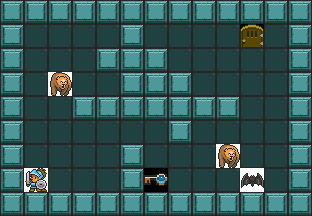}
   }   \end{tabular}
 \label{fig:plotT0}
\end{subfigure}
    \vspace{-1cm}

\end{tabular}

\begin{tabular}{lccccc}
\begin{subfigure}[l]{.08\linewidth}
\centering
    \hspace{\bibindent}\raisebox{\dimexpr 2.5cm-\height}{ GAN~~~~~}\end{subfigure}
&
\begin{subfigure}[b]{.17\linewidth}
\centering
  \begin{tabular}{cc}
 \resizebox{1.0\linewidth}{!}{
  \includegraphics[width=\linewidth]{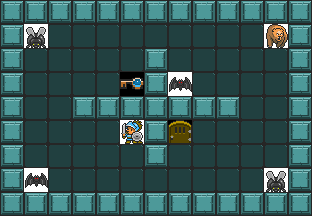}
   }
   \end{tabular}
 \label{fig:plotT0}
\end{subfigure}
\begin{subfigure}[b]{.17\linewidth}
\centering
  \begin{tabular}{cc}
  \resizebox{1.0\linewidth}{!}{
  \includegraphics[width=\linewidth]{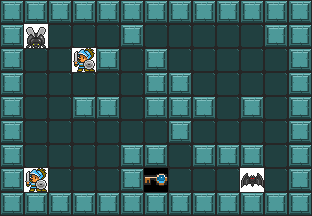}
   }
   \end{tabular}
 \label{fig:plotT0}
\end{subfigure}
\begin{subfigure}[b]{.17\linewidth}
\centering
  \begin{tabular}{cc}
    \resizebox{1.0\linewidth}{!}{
  \includegraphics[width=\linewidth]{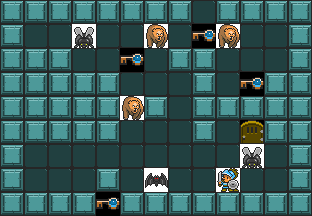}
   }  
   \end{tabular}
 \label{fig:plotT0}
\end{subfigure}
\begin{subfigure}[b]{.17\linewidth}
\centering
  \begin{tabular}{cc}
    \resizebox{1.0\linewidth}{!}{
  \includegraphics[width=\linewidth]{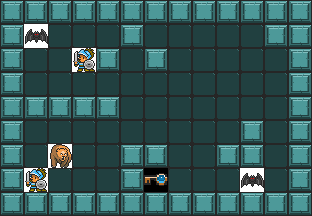}
   }
   \end{tabular}
 \label{fig:plotT0}
\end{subfigure}
\begin{subfigure}[b]{.17\linewidth}
\centering
  \begin{tabular}{cc}
    \resizebox{1.0\linewidth}{!}{
  \includegraphics[width=\linewidth]{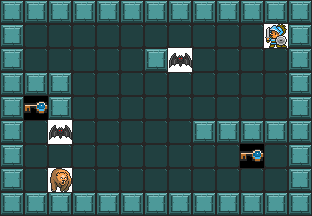}
   }   \end{tabular}
 \label{fig:plotT0}
\end{subfigure}
\vspace{-1cm}

\end{tabular}
\begin{tabular}{lccccc}
\begin{subfigure}[l]{.08\linewidth}
\begin{flushleft}    \hspace{\bibindent}\raisebox{\dimexpr 2.5cm-\height}{  
    GAN+MIP~
    }        \end{flushleft}

    \end{subfigure}
&
\begin{subfigure}[b]{.17\linewidth}
\centering
  \begin{tabular}{cc}
 \resizebox{1.0\linewidth}{!}{
  \includegraphics[width=\linewidth]{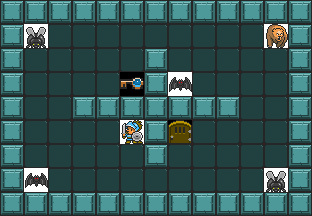}
   }
   \end{tabular}
 \label{fig:plotT0}
\end{subfigure}
\begin{subfigure}[b]{.17\linewidth}
\centering
  \begin{tabular}{cc}
  \resizebox{1.0\linewidth}{!}{
  \includegraphics[width=\linewidth]{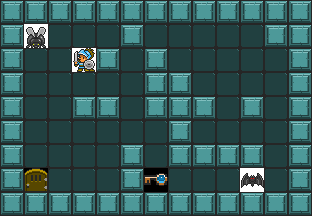}
   }
   \end{tabular}
 \label{fig:plotT0}
\end{subfigure}
\begin{subfigure}[b]{.17\linewidth}
\centering
  \begin{tabular}{cc}
    \resizebox{1.0\linewidth}{!}{
  \includegraphics[width=\linewidth]{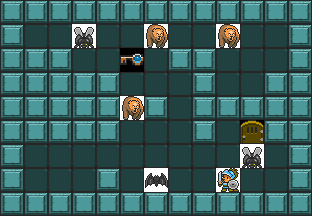}
   }  
   \end{tabular}
 \label{fig:plotT0}
\end{subfigure}
\begin{subfigure}[b]{.17\linewidth}
\centering
  \begin{tabular}{cc}
    \resizebox{1.0\linewidth}{!}{
  \includegraphics[width=\linewidth]{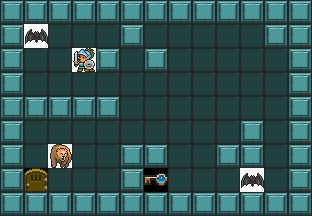}
   }
   \end{tabular}
 \label{fig:plotT0}
\end{subfigure}
\begin{subfigure}[b]{.17\linewidth}
\centering
  \begin{tabular}{cc}
    \resizebox{1.0\linewidth}{!}{
  \includegraphics[width=\linewidth]{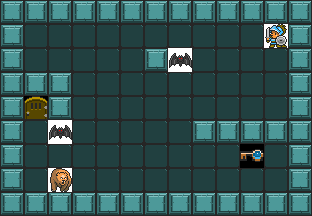}
   }   \end{tabular}
 \label{fig:plotT0}
\end{subfigure}
\vspace{-1cm}
\end{tabular}

\begin{tabular}{lccccc}
\begin{subfigure}[l]{.08\linewidth}
\begin{flushleft}    \hspace{\bibindent}\raisebox{\dimexpr 2.5cm-\height}{  
    MIP-random
    }        \end{flushleft}
    \end{subfigure}
&
\begin{subfigure}[b]{.17\linewidth}
\centering
  \begin{tabular}{cc}
 \resizebox{1.0\linewidth}{!}{
  \includegraphics[width=\linewidth]{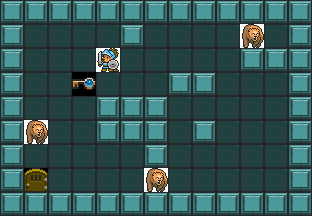}
   }
   \end{tabular}
 \label{fig:plotT0}
\end{subfigure}
\begin{subfigure}[b]{.17\linewidth}
\centering
  \begin{tabular}{cc}
  \resizebox{1.0\linewidth}{!}{
  \includegraphics[width=\linewidth]{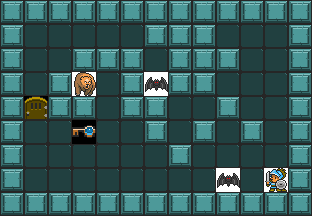}
   }
   \end{tabular}
 \label{fig:plotT0}
\end{subfigure}
\begin{subfigure}[b]{.17\linewidth}
\centering
  \begin{tabular}{cc}
    \resizebox{1.0\linewidth}{!}{
  \includegraphics[width=\linewidth]{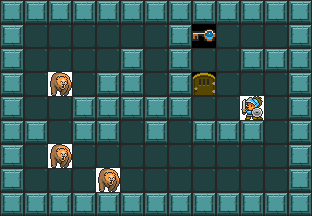}
   }  
   \end{tabular}
 \label{fig:plotT0}
\end{subfigure}
\begin{subfigure}[b]{.17\linewidth}
\centering
  \begin{tabular}{cc}
    \resizebox{1.0\linewidth}{!}{
  \includegraphics[width=\linewidth]{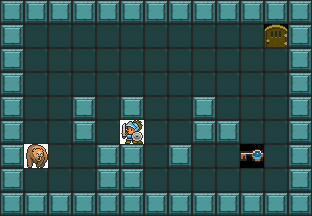}
   }
   \end{tabular}
 \label{fig:plotT0}
\end{subfigure}
\begin{subfigure}[b]{.17\linewidth}
\centering
  \begin{tabular}{cc}
    \resizebox{1.0\linewidth}{!}{
  \includegraphics[width=\linewidth]{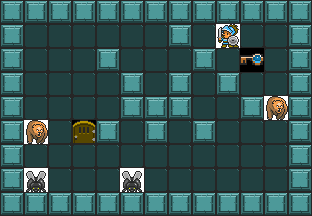}
   }   \end{tabular}
 \label{fig:plotT0}
\end{subfigure}
\vspace{-1cm}
\end{tabular}

\caption{Example generated Zelda levels of different techniques. The GAN+MIP  framework repairs the GAN-generated levels rendering them playable, while capturing the spatial relationships between tiles exhibited in the human-authored levels.}
\label{fig:summary}
\end{figure*}

Our key insight is: 

\begin{quote}
We can create playable levels that are aesthetically similar to human-authored examples, by repairing GAN-generated levels using a mixed-integer linear program with encoded playability constraints. 
\end{quote}

Specifically, the framework first \textit{generates} levels using a GAN trained with human-authored examples. The levels are aesthetically similar to the training levels but may not be playable. We then \textit{repair} the levels using a mixed-integer linear program with playability constraints, which minimizes the number of edits required to render the level playable. A key component of the framework is the edit distance metric, which we cast as a \textit{minimum cost network flow problem}, where we define a separate network for each object type in the level and ``flows'' represent changes between the GAN-generated level and the level generated by the MIP solver.

Fig.~\ref{fig:summary} shows human-authored examples, as well as  levels generated by the GAN and the proposed framework GAN+MIP, with 50 human-authored levels as training data. We additionally include a baseline, MIP (random), where instead of a GAN-generated level, we use as input to the MIP solver a level generated by independently sampling an object type for each tile. This results in levels that are playable, but whose tiles do not exhibit the spatial relationships seen in the human examples.




Our results show that we can generate a diverse range of playable levels that have an aesthetic appeal from a small number of human-authored examples. 

A limitation of our approach is that aesthetic appeal is subjective and we can only use proxy metrics, such as the spatial relationship between tiles. However, we view this as an exciting step towards procedurally generating  aesthetically pleasing levels that satisfy playability constraints.

\section{Problem Description}
We formulate the problem of procedural video game level repair as a discrete
optimization problem. We represent a level as a space graph~\citep{shaker:book16}, where each node in the graph is associated with a region in the game and edges model connections between regions. For instance, in \textit{The Legend of Zelda} video game, the nodes are grid cells representing an object, such as a wall, door or empty space, and the edges connect adjacent cells. 

We then formulate the procedural generation problem as a matching problem, where each node is matched to an available object type. We note that the vast majority of random matchings lead to unplayable levels. In \textit{Zelda}, there is only one player, who should reach a key and a door to exit the level. Walls in the perimeter prevent the player from leaving a confined area. All such constraints need to be satisfied for a level to be playable.


In addition to meeting criteria for playability, video game levels need to be aesthetically pleasing. They must look interesting or engaging to human players. Similarly to aesthetically pleasing images, such considerations are not easily formalized. Here, we formulate this problem as minimizing the number of edits on levels that are sampled from a learned distribution of human-authored training examples.


\section{Domain}
We use a simplified version of \textit{The Legend of Zelda} video game, implemented in the General Video Game Artificial Intelligence (GVGAI) framework~\citep{perez:aaai16, gaina:tog17, perez:tog19}. \textit{Nintendo} first introduced \textit{The Legend of Zelda} in 1986 for the \textit{Nintendo Entertainment System} (\textit{NES}). The main character of the game, Link, must explore dungeons and solve puzzles while simultaneously avoiding enemies within the level. In the GVGAI version, Link must navigate the environment to obtain a key, then proceed to the exit door while avoiding enemies. 

Fig.~\ref{fig:summary}~(top) shows example human-authored \textit{Zelda} levels, with Table~\ref{tab:encoding} showing the different object types. Note that all example human-authored levels are playable. In addition, we observe that the key is typically placed far away from the door, requiring the player to navigate the level, encountering enemies in the way. This is an aesthetic quality that is not a playability constraint, but we wish to replicate in the procedurally generated levels.

\begin{table}\centering
\ra{1.3}
\scalebox{0.8}{
\begin{tabular}{cccccc}\toprule
Wall & Empty & Key & Exit door & Enemy & Player \vspace{0.8em}\\
\includegraphics[scale=0.5]{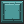} &  \includegraphics[scale=0.5]{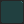}  & \includegraphics[scale=0.75]{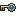}& \includegraphics[scale=0.5]{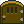}& \includegraphics[scale=0.5]{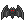}& \includegraphics[scale=0.5]{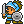}\\
\bottomrule
\end{tabular}
}
\caption{The visualization for the tiles in Zelda.} 
\label{tab:encoding}
\end{table}

\section{Approach}
Our framework implements a \emph{generate-then-repair} approach. For constructing levels adhering to an existing style, we train a generative adversarial network (GAN) using a small number of human-authored examples. The generator network of the GAN samples new levels using random noise as input. The framework passes the GAN-generated level into the objective function of a mixed-integer linear program (MIP) that encodes domain-specific playability constraints. The MIP acts as an editor, minimizing the number of corrections required to transform the GAN generated level into a playable level. This way, we combine the GAN's capability to retain the aesthetics of human-authored examples with the MIP's capability to guarantee formal constraints like those ensuring playability.



\begin{figure}[t!]
\includegraphics[draft=false,width=1.0\columnwidth]{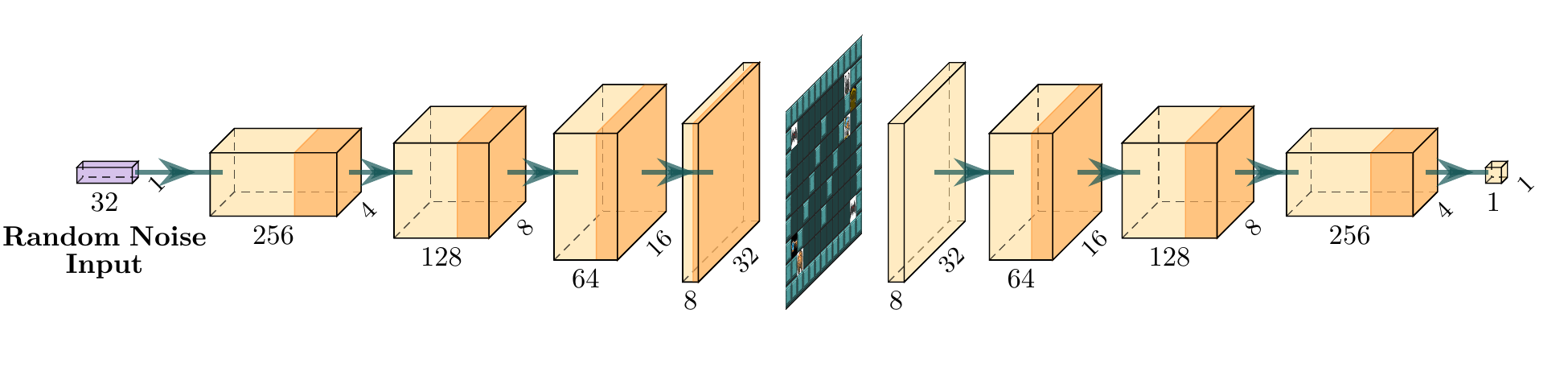}
\caption{DCGAN network for learning the distribution of \textit{Zelda} game levels.}
\label{fig:dcgan}
\end{figure}

\subsection{Deep Convolutional GAN}
We use a Deep Convolutional GAN (DCGAN) with an identical architecture to the DCGAN from \citet{volz:gecco18} (Fig.~\ref{fig:dcgan}), trained with the WGAN algorithm~\citep{martin2017wasserstein}. This model, trained with WGAN, was shown to successfully generate a variety of levels for 
\textit{Super Mario Bros}. 


\subsection{Mixed Integer Linear Program Formulation}
\label{sec:mip_form}
To model reachability as a MIP, let $G(V,E)$ be the space graph, where $V$ is a list of nodes and $E$ is a list of edges. We assume a set of $O$ distinct object types. For each object type $o$, we define a vector of binary variables, $o  \in \{0,1\}^{|V|}$, with each element $o_v$ indicating whether object type $o$ occupies vertex $v$. For example, in \textit{The Legend of Zelda} there are eight object types: wall $w$, empty space $m$, key $k$, door $d$, player $p$ and three types of enemies $e^1$, $e^2$, $e^3$.

\noindent\textbf{Node Uniqueness Constraint.} We require that each node contains exactly one type of object:
\begin{align} \label{eq:limit_objects}
w_v + m_v + k_v + d_v + e^1_v + e^2_v + e^3_v + p_v &= 1,&  \forall v \in V
\end{align}

\noindent\textbf{Reachability Constraints.} Playable levels need to satisfy reachability constraints. For instance, the player needs to be able to reach the key object. We cast the reachability problem as a flow problem~\citep{goldberg:communications14}. For each edge $(u,v) \in E$ we define a non-negative integer variable $f(u,v) \in \mathbb{Z}_{\geq 0}$ representing flow from $u$ to $v$. We define a target set $T \in O$ of object types that need to be reachable by a source set $S \in O$. In Zelda, the door and key need to be reachable by the player, i.e., $T = \{k, d\}$ and $S = \{p\}$. Then, we introduce $f^s_v \in \mathbb{Z}_{\geq 0}$ variables as supplies and $f^t_v \in \{0,1\}$ as demands for each node. We show the network flow equations below, where the equations apply for all nodes $v \in V$:
\begin{align} 
f^s_v &\leq \sum_{x \in S} |V| \cdot x_v &  \label{eq:limit_source}\\
\sum_{x \in T} x_v &= f^t_v & \label{eq:limit_sink}\\
f^s_v + \sum_{u: (u,v) \in E} f(u,v) &= f^t_v + \sum_{u: (v,u) \in E} f(v,u) & \label{eq:flow_conservation}\\
f(u,v) +  \sum_{x_v \in B} |V| \cdot x_v& \leq |V| ~~ ,\forall u: (u,v) \in E \label{eq:blocking_limit}\\
f(u,v), f_v^s  &\in \mathbb{Z}_{\geq 0}\\
f_v^t &\in \{0,1\} &
\end{align}



Eq.~\ref{eq:limit_source} limits supply flow to vertices that have an object in the source set $S$. For instance, in \textit{The Legend of Zelda} we let a node $v$ be connected to the source when $v$ contains the player $p$. Based on Eq.~\ref{eq:limit_objects}, $p_v$ will be 1 and all the other object type variables (e.g., $w_v, m_v$) will be 0. Since $p \in S$, the sum in the right hand side is equal to $|V|$ and $f_v^s$ for that node can take values between 0 and $|V|$. On the other hand, if a node is associated with a tile that contains a non-source node, e.g., a wall, $p_v$ for that node will be 0, forcing the sum on the right side of Eq.~\ref{eq:limit_source} to be 0 and $f_v^s$ will be exactly 0.

Following a similar reasoning, Eq.~\ref{eq:limit_sink} guarantees that node $v$ will generate a unit of demand if it belongs to the target set $T$.
Eq.~\ref{eq:flow_conservation} specifies the flow conservation constraints, which propagate the demands from the target nodes back to the source nodes. The constraints guarantee that all target nodes will be reached by at least one source node. Finally, in Eq.~\ref{eq:blocking_limit} to ensure that no paths cross impassable objects (e.g., walls, door), we define a set of impassable object types $B$ and we block flow leaving nodes assigned to these object types.

\noindent\textbf{Edit Distance Objective.}
Our goal is to minimize the number of edits, that is moving, adding or removing an object type,  that the MIP solver applies to the input level to make it satisfy the playability constraints. We cast this problem as a minimum cost network flow problem, where we generate a  network for every object type. The key intuition is that if a node contains an object in the input level, it is a source node that supplies flow; the supplied flow can be absorbed either by a node with an object of the same type in the MIP solution, which generates a unit of demand, or by a ``waste'' variable that indicates deletion of an object in the original input level. The objective is to minimize the cost of the flow that satisfies supplies and demands.

Similarly to the reachability constraints network, we specify demand variables $f^t_v \in \{0,1\}$ for each node and flow variables  $f(u,v)$ for the edges. We additionally define waste variables $r^t_v \in \mathbb{Z}_{\geq 0}$. 

We let $c_v$ be a constant that is equal to 1 if a node $v$ contained the object type of the network in the input (GAN-generated) level and 0 otherwise. Eq.~\ref{eq:matching_conservation} specifies the flow conservation constraints, while Eq.~\ref{eq:matching_demand_limit} limits demands to locations containing the object type $o$ that the flow network is associated with. The equations apply for all nodes $v \in V$. Eq.~\ref{eq:force_matching} ensures that supplies from initial object locations match demands by the new object locations or by object deletions.

Eq.~\ref{eq:edit_distance_cost} describes the MIP objective to minimize the sum of the costs for the flow network of each object type. The final objective value is computed by summing over all object types (Eq.~\ref{eq:edit_distance_cost}). $r_v^t$ and $f(u,v)$ are different for each object type since they represent flows in different networks. $C_d$ represents the cost for deleting an object and $C_m$ represents the cost for moving an object one tile. Note that we do not model the cost for adding an object as an object must be deleted for an addition to occur. In our experiments, we selected $C_d = 10$ and $C_m = 1$, so that it is much cheaper to move an object one cell than deleting it. 


\begin{align}
c_v + \sum_{u: (u,v) \in E} f(u,v) &= r^t_v + f^t_v +\sum_{u: (v,u) \in E} f(v,u) \label{eq:matching_conservation}\\
f^t_v &\leq o_v \label{eq:matching_demand_limit}
 \\
  \sum_{v \in V} c_v &= \sum_{v \in V} f^t_v + \sum_{v \in V} r^t_v\label{eq:force_matching}
\end{align}


\begin{equation} \label{eq:edit_distance_cost}
\sum_{o \in O} \left( \sum_{v \in V} C_d r_v^t + \sum_{u,v: (u,v) \in E} C_m f(u,v) \right) \; \textrm{(minimize)}
\end{equation}

\noindent\textbf{Domain-specific Constraints.} The above constraints and edit distance objective can generalize across a variety of platform games, such as \textit{The Legend of Zelda} and \textit{Pac-Man}.
However, each game has additional game-specific constraints, which can be easily encoded in the MIP formulation. \textit{The Legend of Zelda} requires that exactly one key, one door, and one player are present in the level, enemies cover less than 60\% of the available space to ensure the level is not too difficult for the player, and the outer perimeter of the level needs to be filled with wall objects. We include these additional constraints in the MIP formulation.

\section{Empirical Evaluation}

\begin{table*}\centering
\centering
\begin{tabular}{@{}rrrrcrrrcrrr@{}}\toprule
\centering
Model & Playable levels & Duplicated levels & Playable and unique levels \\ \midrule
GAN & 24.3\% & 46.9\% & 12.9\% \\
MIP (random) & 100\% & 0\% &  100\%\\
CESAGAN~\citep{torrado:cog20} & 58\% & 37.6\% &  not reported\\
GAN + MIP  & 100\% & 14.9\% & 85.1\% \\
\bottomrule
\end{tabular}
\caption{Percentage of generated playable and unique levels with each technique.}
\label{fig:compare_baseline}
\end{table*}
\begin{figure}[t!]
\centering
\includegraphics[width=0.49\linewidth]{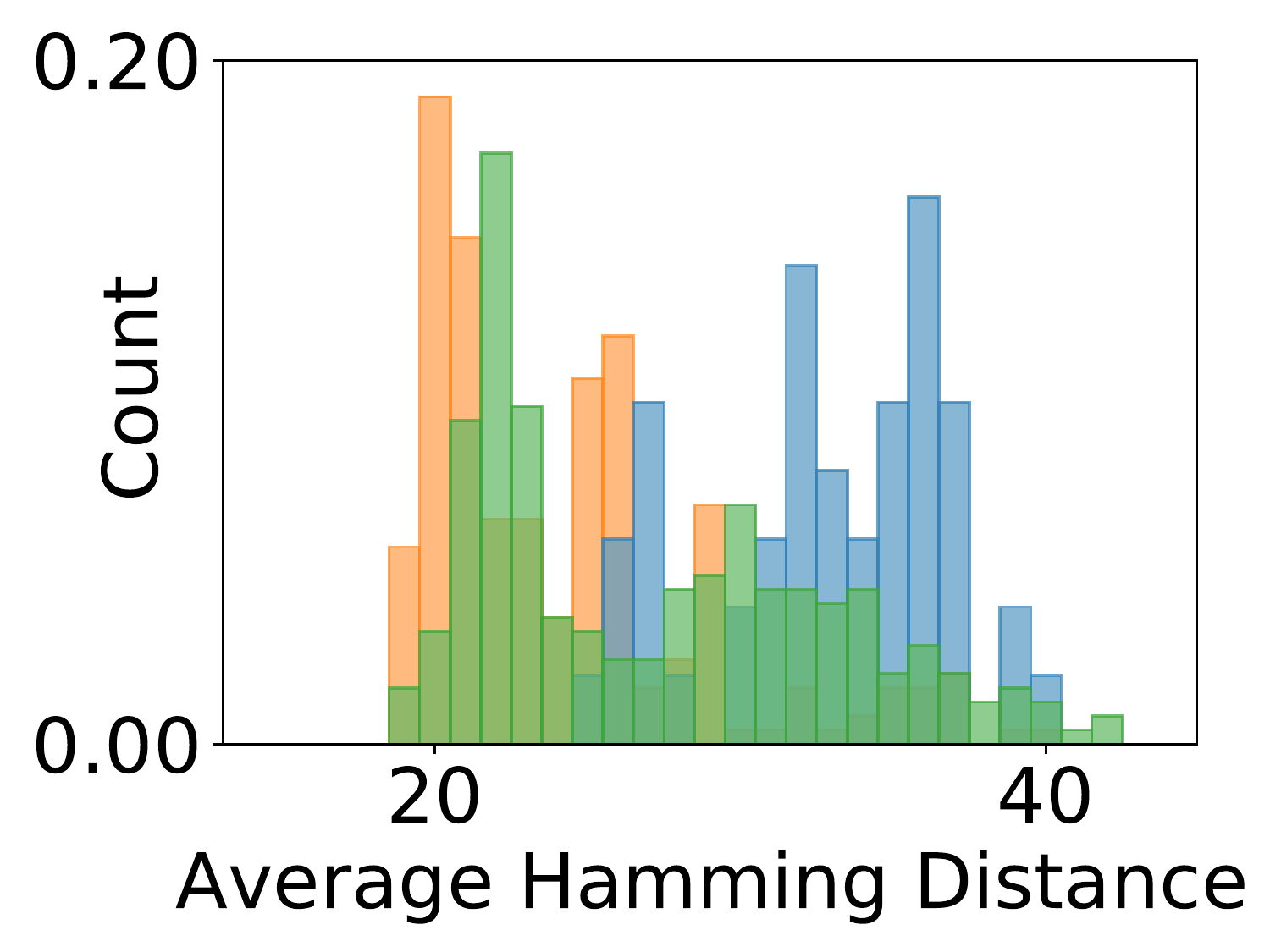}
\includegraphics[width=0.49\linewidth]{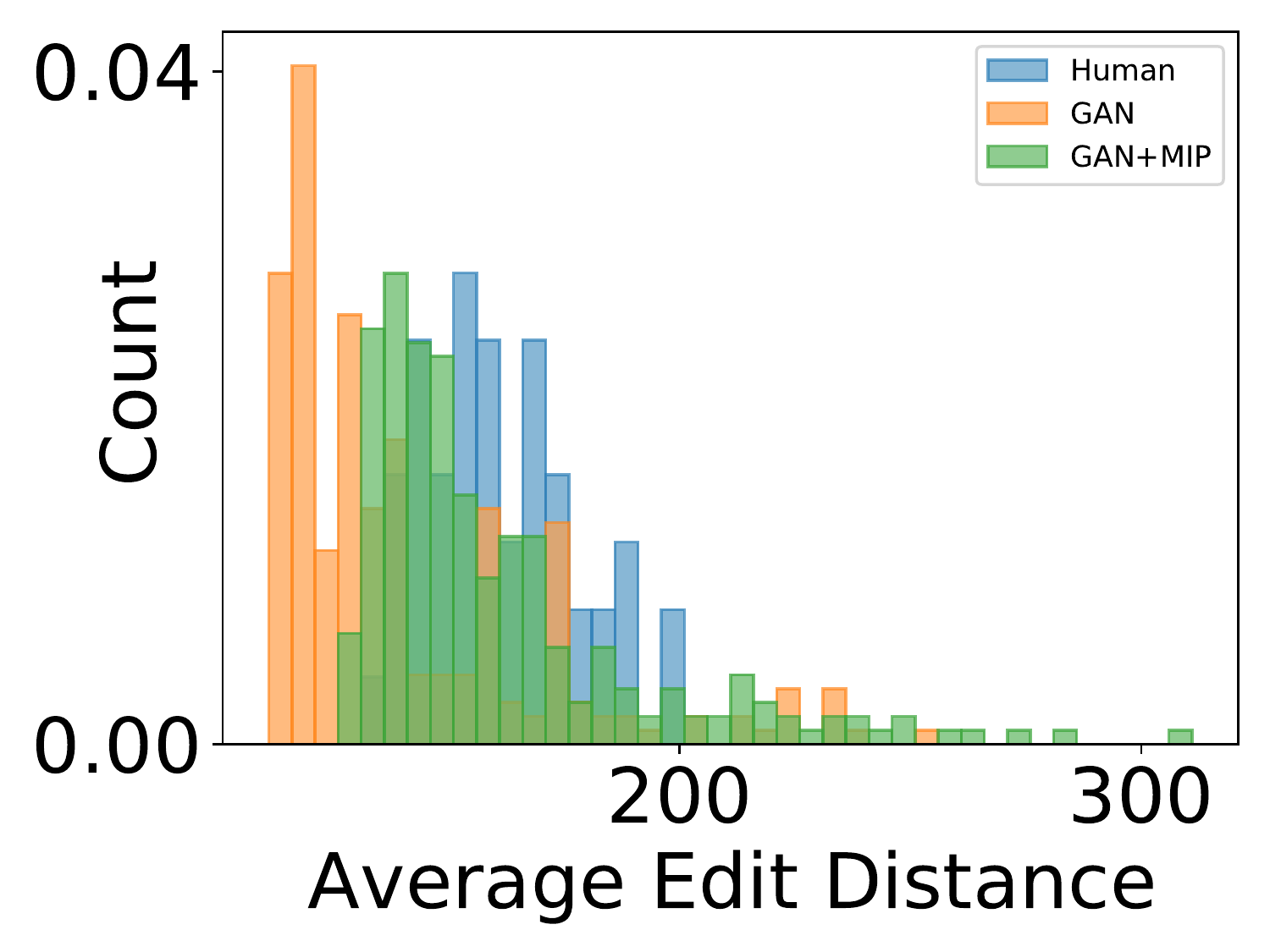}
\caption{The distribution of the average hamming (left) and edit (right) distance between levels from the same set.}
\label{fig:hamming_distance}
\end{figure}

\begin{figure*}[!t]
\centering
\begin{tabular}{llll}
\begin{subfigure}[b]{.23\linewidth}
\centering
\hspace{-2em}
  \includegraphics[width=0.98\linewidth]{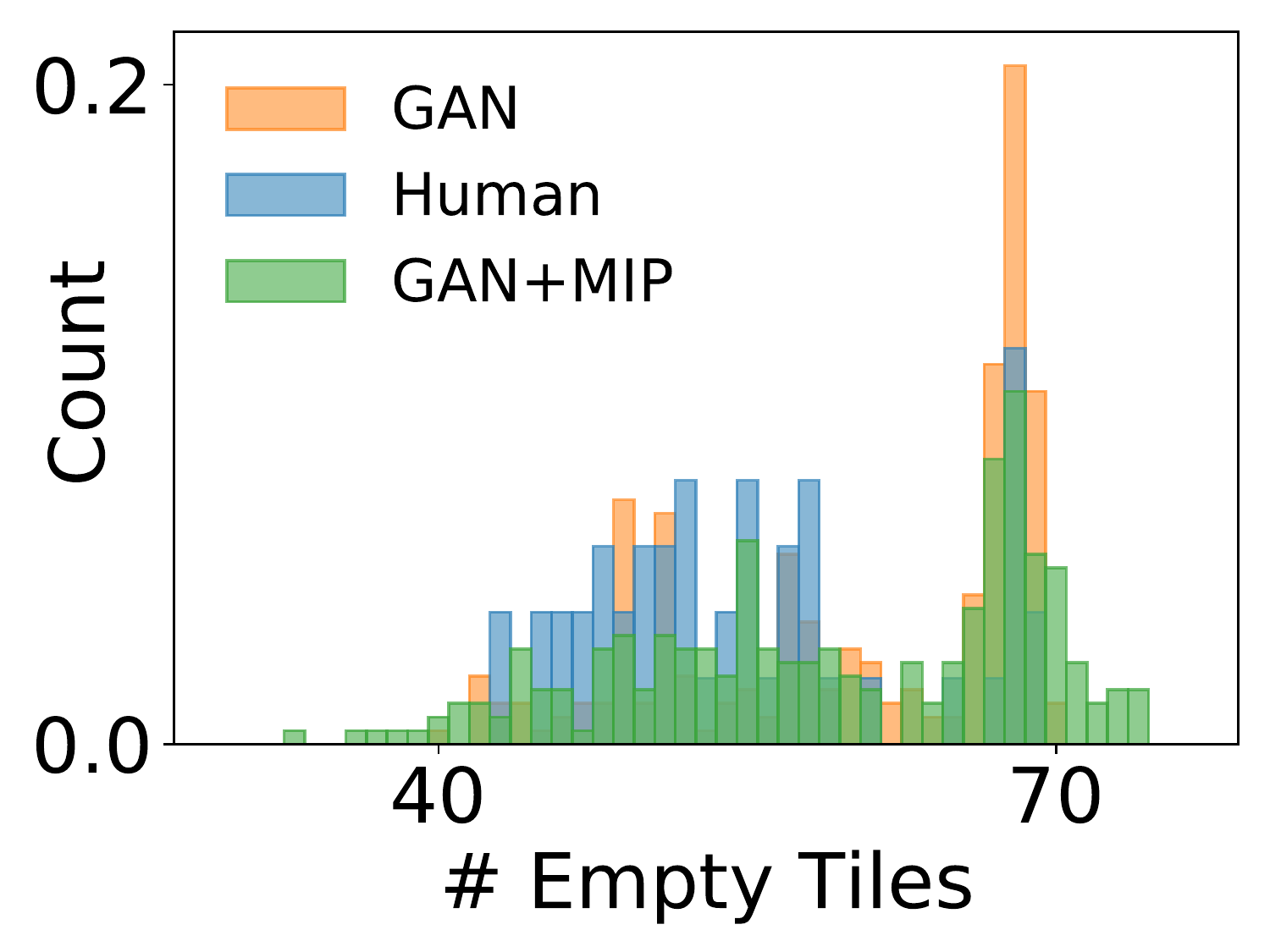}
   \caption{}
 \label{fig:empty}
\end{subfigure}
&
\begin{subfigure}[b]{.23\linewidth}
\centering
  \includegraphics[width=1.0\linewidth]{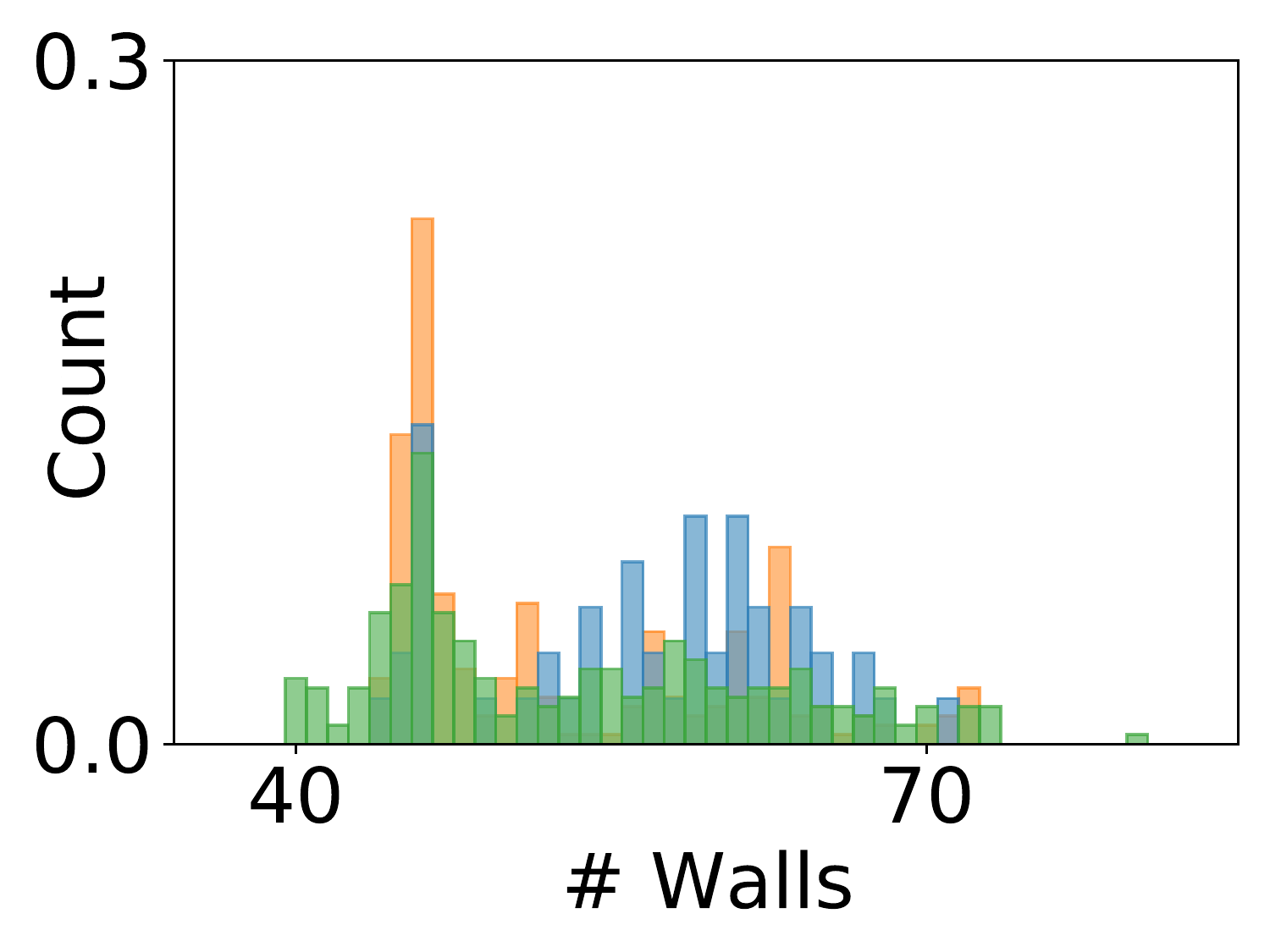}
    \caption{}

 \label{fig:plotT0}
\end{subfigure}
&
\begin{subfigure}[b]{.23\linewidth}
\centering
  \includegraphics[width=0.99\linewidth]{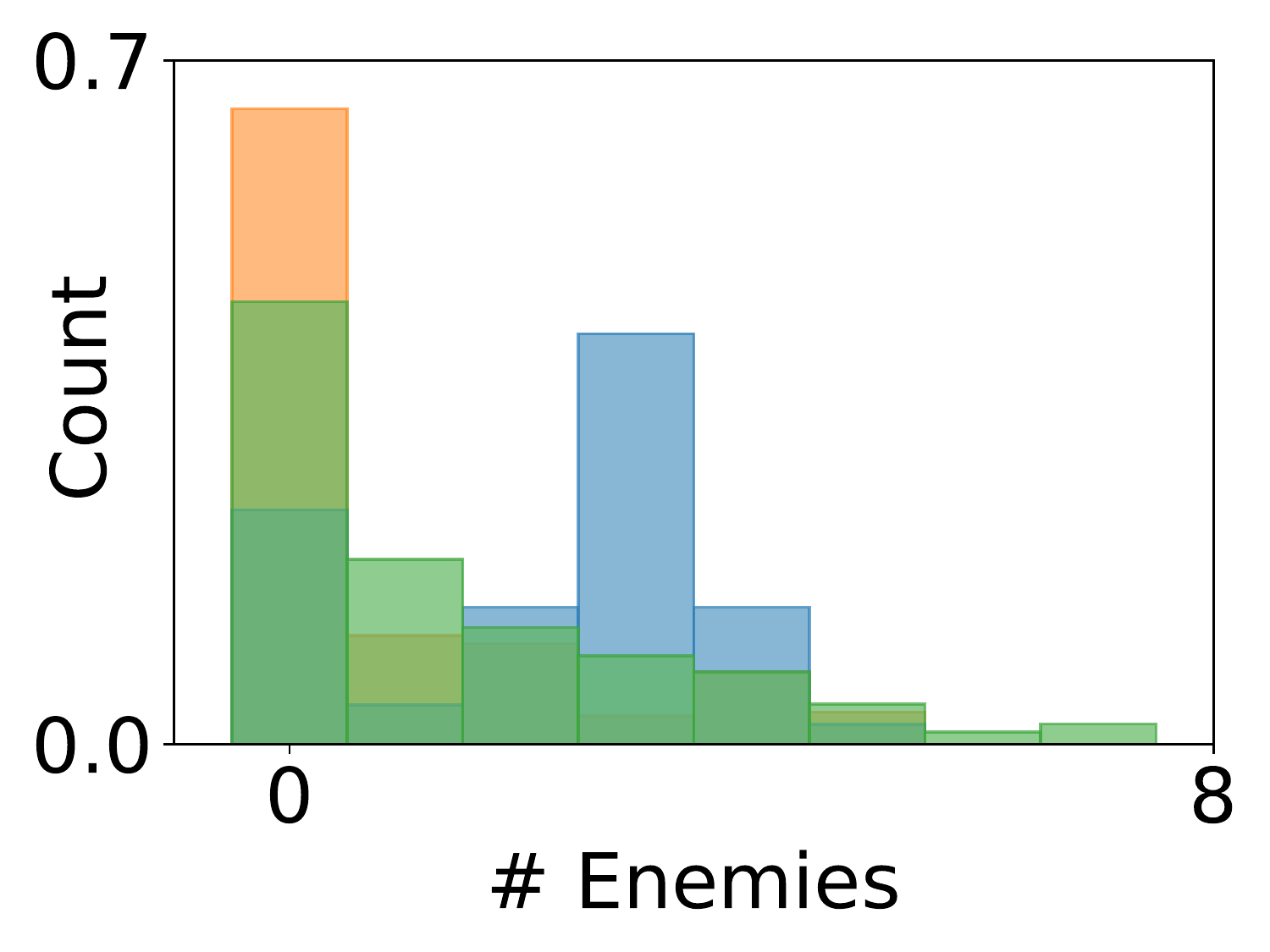}
  \caption{}
 \label{fig:enemies}
\end{subfigure}
&
\begin{subfigure}[b]{.23\linewidth}
\centering
  \includegraphics[width=0.98\linewidth]{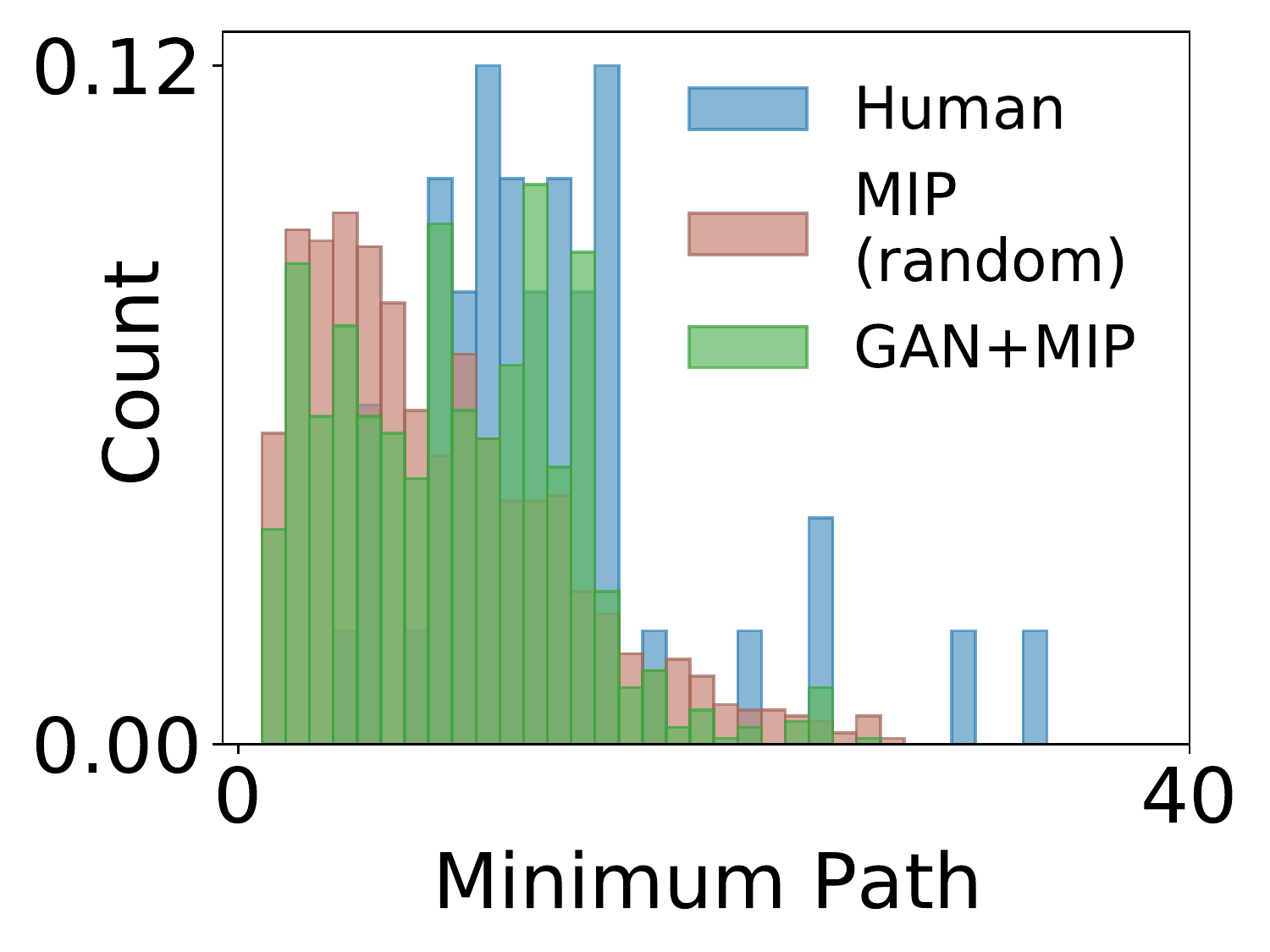}
  \caption{}
 \label{fig:min-path}
\end{subfigure}
\end{tabular}
\caption{(a-c) Distribution of different game tiles for the human examples, the levels generated by the GAN and the levels generated by the GAN and edited with the MIP solver. (d) Distribution of minimum paths from key to door.}
\label{fig:tile_distribution}
\end{figure*}

To evaluate our method we train the GAN on a corpus of 50 human authored levels from \textit{Zelda} for 24,000 epochs. We then generate 1000 levels using the GAN+MIP framework by sampling through Gaussian noise on the generator network and compare against three baselines: levels generated by GAN without the MIP editing and levels generated by the MIP solver by minimizing the edit distance to a randomly generated level and performance results from a recent study~\citep{torrado:cog20}, which uses an adapted self-attention GAN, named CESAGAN. For the randomly generated levels, we sample an object for each tile from a multinomial distribution, where the probabilities match the object frequency counts in the human authored levels. CESAGAN captures non-local dependency between game objects. It was trained on 45 out of the same 50 human-authored levels as the proposed GAN+MIP framework. Note that CESAGAN tries to learn level constraints through bootstrapping and does not require explicitly encoded constraints. 

\noindent\textbf{Diversity of Playable Levels.} We evaluate the generated levels by testing the playability criteria specified in \citet{torrado:cog20}. Levels are also measured for duplicates by counting the number of unique levels produced by the generator and reporting the percentage of additional levels that are duplicates of the unique levels. Table~\ref{fig:compare_baseline} shows our results. 


Results show that most of the GAN-generated levels are not playable, while the proposed framework generates a large number of unique levels that are all playable, since they satisfy the constraints encoded in the MIP. Moreover, we notice that from the duplicate levels generated by the proposed framework, 77\% were generated directly from the GAN. For the remaining duplicate repair levels, each level was stylistically similar; for instance, two distinct levels had a missing boundary tile in different locations, but they were repaired to be identical by filling each missing boundary. On average it took the MIP solver~\citep{ibm_cplex} 0.13 seconds to fix generated levels, where each MIP program consisted of 6858 variables and 2777 constraints.

We further analyze the diversity of the generated levels by using the average Hamming distance metric~\citep{torrado:cog20} (number of different tiles) between each generated level with all other playable levels, as well as the proposed edit distance metric. We randomly selected 243 playable levels out of the 1000 generated ones for the proposed framework, in order to match the number of playable levels generated by the GAN. Fig.~\ref{fig:hamming_distance}  shows that, while the human-authored levels have higher diversity, the proposed framework generates levels that are more diverse than the GAN levels. GAN learns to generate levels from the distribution of human examples, but the generated playable levels are only a small part of the learned distribution. On the other hand, GAN+MIP repairs \textit{all} GAN-generated levels, capturing a larger part of the training distribution. 


We note that the levels generated by MIP-random were all unique and had higher diversity metrics (not shown in the figures), because of the stochastic nature of the random level generation process.


\noindent\textbf{Aesthetic Appeal of Generated Levels.} We assess whether the generated levels are aesthetically similar to the human-authored ones. Following previous work~\citep{torrado:cog20}, we compute the distribution of different game tiles for the human, the GAN and the GAN+MIP levels (Fig.~\ref{fig:empty}-\ref{fig:enemies}). 
We observe that the GAN+MIP matches closer the human distribution, since GAN+MIP repairs all GAN-generated levels, capturing a larger part of the training distribution.\footnote{We exclude the baseline MIP (random) from the analysis, since the objects in the input levels are sampled by the tile distribution of the human-authored levels, therefore the levels follow closely, albeit not exactly because of the editing, the human distribution.}




We also compute the tile-pattern Kullback–Leibler (KL) divergence~\citep{lucas2019tile} between the distribution of the tile patterns of the generated levels and the human-authored levels. We first extract a set of tile patterns by sliding a 2x2 fixed-size window over each level. We then empirically estimate the probability of each pattern for each set of levels and then compute the KL divergence between the two distributions.

The  value of the 243 GAN playable levels is $0.272$. We randomly selected 243 playable levels from the GAN+MIP set, to match the number of the GAN levels. The value was $0.108$, indicating higher similarity to the human examples.

\noindent\textbf{Comparison with MIP (random).} Since the MIP (random) baseline generates unique, playable levels that follow the tile distribution of the human-authored levels, what is the benefit of using a generative adversarial network? 

GANs learn spatial relationships between objects in the level. Specifically, inspection of the human-authored levels in Fig.~\ref{fig:summary} shows that human designers tend to place the key far away from the door, so that the player needs to explore the level before exiting it. This aesthetic quality is lost in many of the random generated levels.

We support this argument in Fig.~\ref{fig:min-path}, which shows the distribution of the length of the minimum paths, computed with a standard A* algorithm~\citep{russell2002artificial}, from the key to the door. We observe that the distribution is shifted towards larger paths for the GAN+MIP levels, compared to the MIP (random) levels. The average length of the minimum path was 8.6 for GAN+MIP, compared to 7.8 for MIP (random) and 12.5 for the human.

\begin{figure}[t!]
\centering
\begin{tabular}{cccc}
\includegraphics[width=0.32\linewidth]{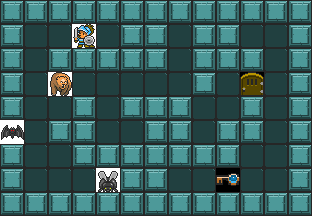}
\includegraphics[width=0.32\linewidth]{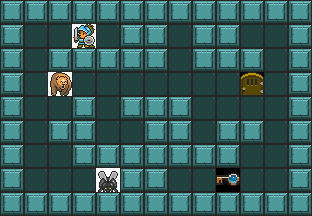}
\includegraphics[width=0.32\linewidth]{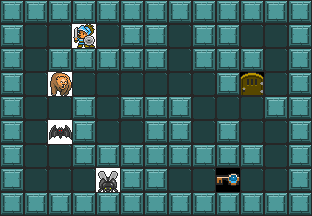}
\end{tabular}
\caption{Edit distance example. (Left) An unplayable generated level by the GAN network. (Center) The output of the MIP solver that minimizes the Hamming distance to the input level. (Right) The output of the MIP solver that minimizes the edit distance.}
\label{fig:edit_distance_example}
\end{figure}

\section{Edit Distance Objective}

A simpler optimization objective for the MIP program would be  the Hamming distance (number of different tiles) rather than the edit distance between the input and the generated level. However, this metric does not capture the spatial relationships between tiles. For instance, in Fig.~\ref{fig:edit_distance_example} minimizing the Hamming distance results in replacing the enemy on the left hand side with the wall. In the edit distance metric of Eq.~\ref{eq:edit_distance_cost}, the cost of deleting an enemy ($C_d = 10) $ is larger than the cost of moving it ($C_m = 1$), therefore the solver chooses to move the enemy and a neighboring wall tile so that they exchange positions. This preserves the level topology by retaining the third enemy.

\section{End-to-End Training}
We explore integrating playability constraints as an additional layer in the GAN network. Recent advances in differential optimization~\citep{wilder:aaai19,ferber:arxiv19} have allowed integrating discrete optimization problems into deep learning models trained with gradient descent. We include a differentiable MIP program~\citep{ferber:arxiv19} as an additional layer in the GAN network and train the GAN-MIP in an end-to-end manner by passing to the discriminator the levels generated by the generator after they are repaired by the MIP solver.


Training the network with all playability constraints is computationally expensive. On the other hand, we observed that the main reason that GAN-generated levels are rendered unplayable is the violation of numerical constraints, such as number of players, doors and keys, which matches results from previous work~\citep{torrado:cog20}. For computational efficiency, we encoded only these constraints in the MIP program, used the simpler Hamming distance objective and applied LP-relaxation~\citep{wilder:aaai19}, which significantly sped up the training process. 

We trained the resulting network for 5000 epochs, which lasted 55 hours on an Intel Core i7-8700K 3.7GHz processor. The generator network generated 747 unique and playable game levels out of 1000, which  is a significant improvement to the initial GAN model. The average length of the minimum path from the key to the door was 10.1, and the KL divergence between the tile distribution of the generated levels and the human-authored levels was 0.055. These  preliminary results indicate the promise of integrating MIP constraints in the GAN training process. 

\section{Beyond \textit{Zelda} Levels}
\begin{figure}
    \centering
    \begin{tabular}{cc}
    \includegraphics[width=0.45\linewidth]{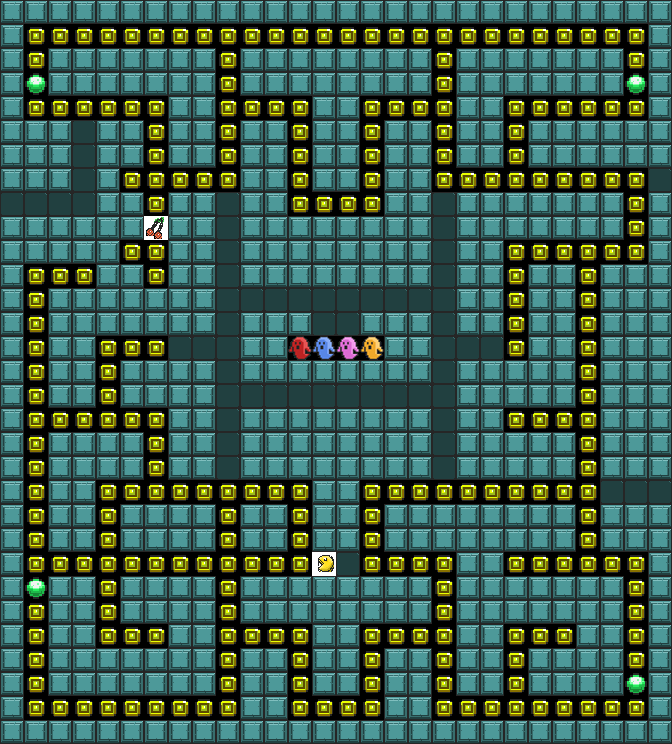}&
     \includegraphics[width=0.45\linewidth]{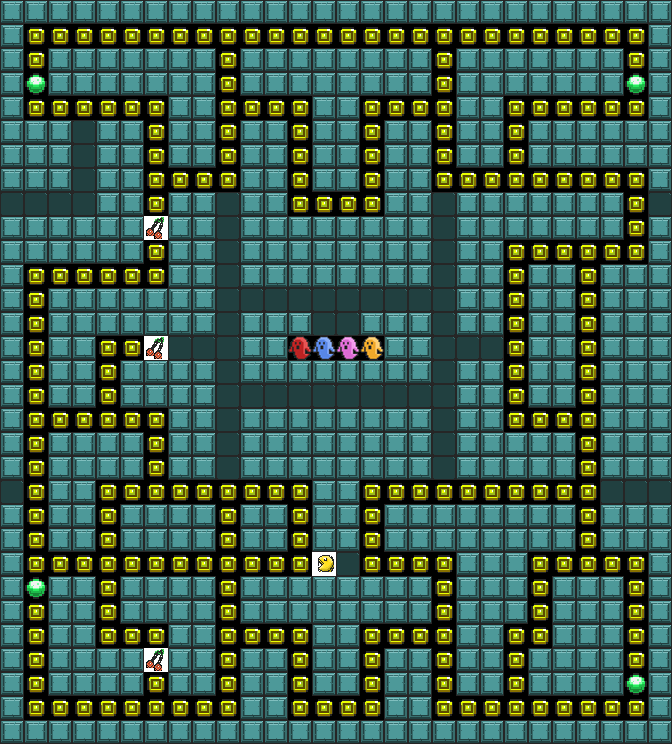}
     \end{tabular}
    \caption{GAN-generated \textit{Pac-Man} level (left) and the same level repaired by the MIP solver (right).}
    \label{fig:Pacman}
\end{figure}

The proposed framework applies also to other domains, such as \textit{Pac-Man} game levels. Modeling playability constraints in \textit{Pac-Man} required small modifications to the \textit{Zelda} MIP. First, the dynamics of \textit{Pac-Man} allow for the character to leave the left side of the screen to enter the right. We modify the space graph to have additional edges between the left column's nodes and the right column's nodes. Similarly, we add edges between the top row's nodes and the bottom row's nodes. \textit{Pac-Man} levels are required to have no dead ends. To model this in the MIP, we require that all \textit{free space} objects are adjacent to at least two neighbors (in graph theory terms: the node is not a leaf). Fig.~\ref{fig:Pacman} (left) shows an example level generated by the DCGAN, trained over 45 human-edited examples. The level has dead-ends and the wraparound on the edges of the screen is incorrect. Fig.~\ref{fig:Pacman} (right) shows the repaired level by a MIP program that minimizes the edit distance to the previous level while satisfying the no-deadend and wraparound constraints, in addition to requiring the player to be able to reach all pellets and enemies in the level. We note that the MIP solver~\citep{ibm_cplex} took 3.22 seconds on average to fix the generated levels. The repairs took longer than the \textit{Zelda} domain as the \textit{Pac-Man} domain has more constraints and a larger space graph. Each MIP program for \textit{Pac-Man} consisted of 65352 variables and 26252 constraints, compared to 6858 variables and 2777 constraints for \textit{Zelda}.

\section{Generality and Limitations}
In terms of expressive power, any satisfiability (SAT) program can be modelled as a MIP. However, not all constraint programs (CP) can be modelled as a mixed integer linear program due to the linearity requirements. What MIPs lack in generality they make up in performance of the solver. Modern MIP solvers can solve programs with millions of variables and constraints, in general orders of magnitude larger than general CP solvers. Moreover, problems like flow have good LP relaxations, allowing for the subprogram to be solved in polynomial time.

While further research is needed in this direction, we argue that many of the PCG methods currently being modelled as CP can be modelled as MIPs, allowing for larger levels to be solved by more efficient solvers. For example, in \textit{Zelda} we may want to generate levels with two doors, where the player must not be able to reach the second door without passing through the first door. The problem can be modelled as an $(s,t)$-cut problem, which (like flow) can be modelled as a linear program. Our goal is to require a 0 cost cut between the player and second door. We link the player to the source vertex and the second door to the sink vertex. Variables $d(u,v) \in \mathbb{Z}_{\geq 0}$ represent a decision variable marking whether edge $(u,v)$ should be in the cut. Then we add constraints that enforce $d(u,v)$ to be zero for edges leaving free space nodes forcing the cut to only cut edges leaving blocking nodes (i.e. the walls or door one). 

More complex reachability constraints can also be modelled. \citet{horswill:aiide2020} demonstrated a constraint programming method for modeling path constraints between fixed points in a level. The CP required that a player could reach enough resources (e.g. ammo, armor) to complete the level. To model this constraint as a MIP we could model the reachability as flow with costs rather than flow, where costs represent obtaining resources. Then we require that the cost of the path from source to sink is constrained within a specific range. Finally, aesthetic constraints like symmetry can be modelled by adding constraints for each pair of nodes on opposite sides of a level requiring each object assignment variable $o_v$ equal it's counterpart across the line of symmetry.

While MIPs can be used as powerful modeling tools, applying these techniques to level repair requires expertise in MIP modeling. Common problems, such as reachability, can be modelled from their graph-based representation and converted to MIP constraints within our framework to ease the burden of designers. Further research is needed to model games with complex physics for character movement. Examples include platformer games like \textit{Mario} and \textit{Sonic the Hedgehog}. For these games heuristic modeling, such as requiring two platforms are close enough for the character to jump between platforms, is needed. 


\section{Related Work}

\citet{volz:gecco18} and \citet{giacomello:gem18} presented the first works applying GANs to the problem of automatic video game level generation. To assure that generated levels satisfy specific criteria, \citet{volz:gecco18} adapted latent variable evolution (LVE) from \citet{bontrager:icbtas18} to search the generative space induced by the generator network with CMA-ES~\citep{hansen:ec01, hansen:arxiv16}. \citet{giacomello:gem18} assured that levels met necessary criteria through generate-and-test methodology. Later work by  \citet{torrado:cog20} showed that GANs can fail to capture logical constraints of the video game levels and invented a bootstrapping method that incorporates generated playable levels back into a continually expanding training data set. \citet{snodgrass:ijcai17} introduced a markov chain method for guaranteeing aspects of the layout in \textit{Mario}. A generate-and-test method guaranteed reachability constraints by running A* agents on generated levels. As an alternative to GANs, WaveFunctionCollapse~\citep{karth:fdg17} generates levels that match example data by compiling the data into constraints that can be used to generate levels of similar visual style. However, note that such approaches cannot model complex playability constraints specified by a user.

Several authors present machine learning approaches to satisfying level constraints. However, these approaches make no guarantees about successfully generating or repairing levels to satisfy all constraints. \citet{karth:fdg19} proposed using discriminator networks, trained on positive and negative examples, to guide WaveFunctionCollapse towards playable levels. \citet{jain:iccc16} proposed using autoencoders as a repair method for broken levels by passing broken components through an autoencoder trained on valid levels. However, each of these approaches assumes that the deep learning model can capture complex logical constraints, which often require specialized models~\citep{wang:icml19}.

As an alternative to procedural content generation via machine learning (PCGML), there exist several methods for declarative modeling of procedurally generated content. \citet{smith:coai11} presented a declarative method of PCG allowing the generative space of possible levels to be encoded as an answer set program. \citet{smith:tanagra11} presented an interactive constraint programming method for allowing humans to coauthor Mario levels with the generative method. \citet{horswill:aiide2020} presented a constraint programming method for modeling path constraints between two fixed points in a space graph. Their method modelled paths as a system of linear equations, which we note is equivalent to flow conservation constraints used in our method (linear programming is a generalization of linear systems of equations). Our method can therefore be thought of as a generalization of their approach that allows for endpoints of the path to be placed dynamically and can model reachability across a set of object types. Constraints can also be encoded in a multi-objective evolutionary algorithm, either as part of the objective function, or separately, e.g., by dividing a population of solutions into feasible and infeasible individuals~\citep{kimbrough2008feasible,sorenson2010towards}.  The two-population constraint handling approach can also be used together with quality diversity algorithms ~\citep{liapis2015constrained, khalifa2018talakat} to generate a diverse range of levels.

Our work benefits from recent advancements combining machine learning with traditional optimization methods. Recent works have introduced quadratic programming~\citep{amos:icml17}, linear programming~\citep{wilder:aaai19}, and satisfiability~\citep{wang:icml19} solvers as layers of deep neural networks. Related methods incorporate submodular optimization~\citep{wilder:aaai19} as a layer in a neural network.

\section{Conclusions}
\noindent\textbf{Limitations.} Our work is limited in many ways. Ensuring reachability constraints requires the agent dynamics be modeled as a finite graph, with vertices representing discrete regions in the level and edges indicating neighboring vertices. Levels that satisfy playability constraints may still be unplayable in practice, for instance if difficulty is too high. Additionally, the objective metrics for aesthetic similarity to human examples are only approximate metrics. Evaluating playability and aesthetic appeal with user studies is a natural extension of this work. 

\noindent\textbf{Implications.} We have presented a \textit{generate-then-repair} framework for constructing levels using models trained over human-authored examples and repairing the levels with a mixed-integer linear program. Our editing method is agnostic to how levels are generated; we are excited to explore the limits of our approach by repairing different types of procedurally generated content (levels, objects, enemies) that need to satisfy explicitly defined constraints.

\section{Acknowledgements}
The authors would like to thank Per Josefsen and Nicola Zaltron, who authored the 45 human-designed \textit{Zelda} levels, and Ahmed Khalifa for his helpful comments about the CESAGAN approach.

\medskip

\small

\bibliographystyle{aaai}
\bibliography{bibliography}

\end{document}